\PassOptionsToPackage{hyphens}{url}
\documentclass[letterpaper]{article} % DO NOT CHANGE THIS
\usepackage{aaai2026}  % DO NOT CHANGE THIS
\usepackage{times}  % DO NOT CHANGE THIS
\usepackage{helvet}  % DO NOT CHANGE THIS
\usepackage{courier}  % DO NOT CHANGE THIS
\usepackage[hyphens]{url}  % DO NOT CHANGE THIS
\usepackage{graphicx} % DO NOT CHANGE THIS
\usepackage{natbib}  % DO NOT CHANGE THIS AND DO NOT ADD ANY OPTIONS TO IT
\usepackage{caption} % DO NOT CHANGE THIS AND DO NOT ADD ANY OPTIONS TO IT
\usepackage{algorithm}
\usepackage{algorithmic}
\usepackage{amsthm}

\usepackage{newfloat}
\usepackage{listings}
\DeclareCaptionStyle{ruled}{labelfont=normalfont,labelsep=colon,strut=off} % DO NOT CHANGE THIS
\floatstyle{ruled}
\newfloat{listing}{tb}{lst}{}
\floatname{listing}{Listing}
\title{DeepProofLog: Efficient Proving in Deep Stochastic Logic Programs}
\author{
    Ying Jiao\footnote{Equal first authorship}\textsuperscript{\rm 1},
    Rodrigo Castellano Ontiveros\footnotemark[1]\textsuperscript{\rm 2},
    Luc De Raedt\textsuperscript{\rm 1,3},
    Marco Gori\textsuperscript{\rm 2},
    Francesco Giannini\textsuperscript{\rm 4,5},
    Michelangelo Diligenti\textsuperscript{\rm 2},
    Giuseppe Marra\textsuperscript{\rm 1}
}
\affiliations{
    \textsuperscript{\rm 1}KU Leuven, Belgium\\
    \textsuperscript{\rm 2} University of Siena, Italy \\
    \textsuperscript{\rm 3} Örebro University, Sweden \\
    \textsuperscript{\rm 4} Scuola Normale Superiore, Italy \\
   \textsuperscript{\rm 5} University of Pisa, Italy \\
    firstname.lastname@kuleuven.be, firstname.lastname@unisi.it, francesco.giannini@sns.it
}

\usepackage{bibentry}
\usepackage{amsmath}
\usepackage{amsfonts}
\usepackage{booktabs}
\usepackage{adjustbox}
\usepackage{forest}
\usepackage{multirow}
\newcommand{\our}{DeepProofLog}
\newcommand{\acr}{DPrL}
\newcommand{\mc}{\mathcal }
\newtheorem{definition}{Definition}
\newtheorem{proposition}{Proposition}

\begin{document}

\maketitle

\begin{abstract}
Neurosymbolic (NeSy) AI combines neural architectures and symbolic reasoning to improve accuracy, interpretability, and generalization. While logic inference on top of subsymbolic modules has been shown to effectively guarantee these properties, this often comes at the cost of reduced scalability, which can severely limit the usability of NeSy models. 
This paper introduces \our{} (\acr), a novel NeSy system based on stochastic logic programs, which addresses the scalability limitations of previous methods. \acr{} parameterizes all derivation steps with neural networks, allowing efficient neural guidance over the proving system. Additionally, we establish a formal mapping between the resolution process of our deep stochastic logic programs and Markov Decision Processes, enabling the application of dynamic programming and reinforcement learning techniques for efficient inference and learning. This theoretical connection improves scalability for complex proof spaces and large knowledge bases. Our experiments on standard NeSy benchmarks and knowledge graph reasoning tasks demonstrate that \acr{} outperforms existing state-of-the-art NeSy systems, advancing scalability to larger and more complex settings than previously possible. 
\end{abstract}

\begin{links}
    \link{Code}{https://github.com/DeepProofLog/DPrL-AAAI}
    %\link{Datasets}{https://aaai.org/example/datasets}
%     \link{Extended version}{http://arxiv.org/abs/2511.08581}
\end{links}

\section{Introduction}
Neurosymbolic (NeSy) AI represents a promising paradigm that seeks to bridge the gap between data-driven neural architectures and the structured reasoning capabilities of symbolic methods to obtain improved accuracy and enhanced interpretability.
A key research direction within NeSy focuses on integrating probability with logic programming, enabling principled learning and inference under uncertainty \cite{manhaeve2018deepproblog, dai2019bridging, marra2021neural, pryor2022neupsl, maene2023soft, yang2023neurasp, de2023neural}. These systems typically perform inference in two stages: (1) apply symbolic proving to find proofs for a given query, (2) rely on probabilistic inference over these proofs to compute the probability that the query holds. This process employs the \textit{possible world semantics} \cite{de2007problog, manhaeve2018deepproblog}, which computes the probability of a query as the sum of the probabilities of those possible worlds that have at least one proof for the query. 
Regrettably, this probabilistic inference task corresponds to the weighted model counting problem, which is \#P-hard \cite{roth1996hardness}. 

To alleviate the scalability challenges in probabilistic-logic NeSy systems \cite{wang2013programming,maene2024hardness, marra2024statistical,jiao2024valid}, most existing studies focus on scaling the probabilistic inference phase, assuming that derivations of queries can be computed tractably, including sampling approaches \cite{verreet2024explain} and variational approaches \cite{abboud2020learning,dos2021neural,van2023nesi}.
Although these methods represent progress in the scalability of NeSy, their applicability to large-scale relational settings remains out of reach.
This limitation arises from two primary factors. First, many solutions are still ad hoc, tailored to specific tasks, and require substantial manual effort to adapt to new domains. Second, and more fundamentally, these methods do not address the computational cost of deriving the proofs.

\begin{figure*}[t]
  \centering
\includegraphics[width=0.99\textwidth]{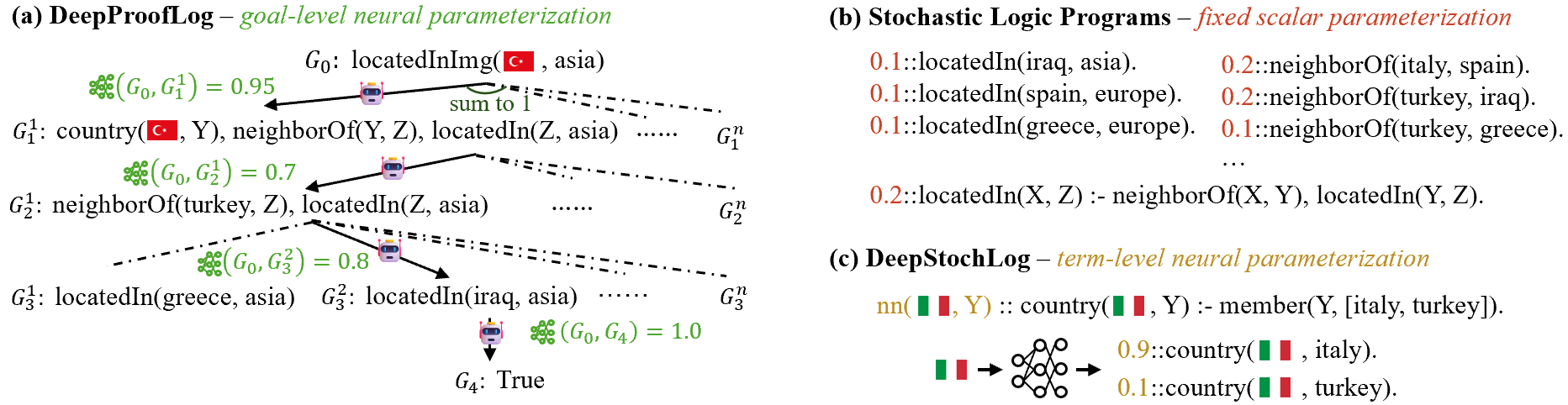}
\caption{(a) \our{} uses goal-level parameterization: a neural policy conditioned on the goal to guide proving at each resolution step. In comparison, (b) SLPs assign scalar weights to clauses for probabilistic proof search without input adaptation,
and (c) DeepStochLog uses neural nets conditioned on subsymbolic terms in the query, requiring a fixed output domain.}
  \label{fig:param_compare}
\end{figure*}

This paper introduces \our{} (\acr{}), a novel system that takes a different perspective to significantly advance NeSy scalability (see Figure \ref{fig:param_compare}). Instead of relying on possible world semantics, our work focuses on the semantics of Stochastic Logic Programs (SLPs) \cite{cussens2001parameter}, which define a probability distribution over \emph{derivations} rather than over possible worlds. 
The probability associated with a clause in SLPs
reflects its likelihood of being used in a successful query derivation rather than its truth in a possible world.
From this perspective, SLPs do not directly model uncertainty about the world, but rather the uncertainty in the behavior of a prover operating within that world. This view offers an agent-based interpretation of inference and learning in NeSy, which, to the best of our knowledge, has not been previously explored.
Furthermore, we define a formal correspondence between learning and inference in (deep) SLPs and the problem of learning optimal policies in Markov Decision Processes (MDPs) \cite{feinberg2012handbook}. In our formulation, each derivation corresponds to a trajectory in the MDP, and each proving step is parameterized by a neural network that works as a policy by providing continuous and goal-directed guidance for selecting the next action (i.e. a rule application or variable substitution). \textcolor{red}{}This connection allows us to draw upon the rich body of established MDP solution methods, which have proven effective in a variety of complex domains, such as game playing and robotics, to tackle scenarios that have traditionally been intractable for NeSy systems.
In particular, when the symbolic goal space is computationally manageable, we show that dynamic programming techniques can be used for exact inference with enhanced efficiency.
Conversely, when the complete goal space becomes intractable, we apply deep reinforcement learning (RL) to approximate the optimal policy through learning from sampled derivations. 

While automated theorem proving (ATP) can also be framed as an MDP and solved with RL, NeSy reasoning differs conceptually. ATP aims to find any valid proof, whereas NeSy focuses on probabilistic logics, weighing multiple proofs and assigning low probabilities to those inconsistent with subsymbolic signals such as images or embeddings. Consequently, DPrL must both identify plausible proofs and ensure their probabilities align with these subsymbolic signals—this alignment is the central challenge we address to improve efficiency in NeSy learning.

To our knowledge, the only other NeSy system designed as a neural extension of SLPs is DeepStochLog \cite{winters2022deepstochlog}. It is based on stochastic definite clause grammars, a type of SLPs, and introduces neural grammar rules. While DeepStochLog represents a step forward, it still faces challenges with scalability and expressiveness compared to \acr{}. 
A more technical comparison between our method and DeepStochLog is reported in Section~\ref{sec:rel_work}.
% The paper is organized as follows. Section~\ref{sec:prel} recalls the basic notions underlying \acr{}. Our model architecture is described in Section~\ref{sec:method}, while Section~\ref{sec:exp} presents the experimental analysis. Finally, Section~\ref{sec:rel_work} describes related work and Section~\ref{sec:conc} concludes.

\noindent\textbf{Contributions.} (i) We introduce \our{} (\acr{}), a novel deep SLP model that leverages neural parameterization at each proof step to provide continuous, goal-directed guidance during reasoning. (ii) We propose a theoretical mapping between derivation-based probabilistic logic reasoning and MDPs, enabling the use of powerful MDP solution techniques for efficient learning and inference in \acr{}.  (iii) Experiments on NeSy benchmarks and knowledge graph completion tasks empirically validate the effectiveness of our approach, achieving improved task performance, efficiency, and scalability over other NeSy systems.

\section{Preliminaries} \label{sec:prel}

\subsection{Logic Programming}
Following \cite{flach1994simply,lloyd2012foundations}, we summarize the basic concepts of logic programming. 
A term is either a constant, a variable, or a compound term of the form $f(t_1, ..., t_n)$, where $f$ is a functor of arity $n$ and each $t_i$ is a term. An expression $r(t_1, ..., t_n)$,  where $r$ is a predicate and each $t_i$ is a term, is called an atom. For example, $\textit{locatedIn}(X,\textit{europe})$ is an atom where $\textit{europe}$ denotes a constant and $\textit{locatedIn}$ a binary predicate. 
A definite clause\footnote{We will omit ``definite'' as all clauses in this paper are definite.} is of the form $H \leftarrow B_1, ..., B_m$, where $H$ is the \textit{head} atom and $B_1, \dots, B_m$ form the \emph{body} atoms. 
A set of definite clauses $\mc{P}$ is called a \textit{definite logic program}. 
A clause of the form ${}
\leftarrow A_1, \dots, A_n$ is called a \textit{goal} and represents a query to the program. A \textit{substitution} $\theta$ such as $\{X/ italy \}$ is a mapping from variables to terms. A substitution $\theta$ is a \textit{unifier} of two atoms (same for terms) $A_1$ and $A_2$ if $A_1\theta$ (application of $\theta$ to variables of $A_1$) = $A_2\theta$. For instance, $\theta=\{X/\textit{italy},Y/\textit{europe}\}$ unifies $A_1=\textit{locatedIn}(X,\textit{europe})$ and $A_2=\textit{locatedIn}(\textit{italy},Y)$, as $A_1\theta=A_2\theta=\textit{locatedIn}(\textit{italy},\textit{europe})$. A substitution $\theta$ is the Most General Unifier (MGU) if any other unifier can be obtained by further instantiating $\theta$—i.e., it makes the minimal necessary variable substitutions.

Given a logic program $\mathcal{P}$, a Selective Linear Definite (SLD) \emph{derivation} of an initial goal $G_0$ is defined as a (finite or infinite) sequence of goals $G_0 \vdash G_1 \vdash G_2 ...$ where for each $i \geq 0$ we have: i) $G_i = {} \leftarrow A_1^i, \dots, A_n^i$, ii) $C_i = H_i \leftarrow B_1^i, \dots, B_m^i$ is a clause in $\mathcal{P}$, with $A_1^i$ and $H_i$ unifiable with MGU $\theta_i$, iii)  $G_{i+1}$ is constructed from $G_i$ by replacing $A_1^i$ with the body of $C_i$, and by applying the substitution $\theta_i$ to each atom in the resulting goal, i.e.
  $
    G_{i+1} = ({} \leftarrow B_1^i, \dots, B_m^i, A_2^i, \dots, A_n^i)\theta_i
  $.
We denote this resolution step by $G_{i+1} = \operatorname{res}(G_i, C_i, \theta_i)$.
The process proceeds by repeatedly applying the above resolution steps until the goal reduces to the empty/True goal (successful derivation), or no further resolution is possible, thus the goal is False (failure derivation).

\subsection{Stochastic SLD Resolution}
Stochastic SLD resolution defines a probability distribution over derivations. This framework captures uncertainty in the resolution process and enables the learning of probabilistic models from data. As special cases, both SLPs and DeepStochLog can be derived from this approach. Their relations are illustrated in Figure~\ref{fig:param_compare}.

\begin{definition}\textbf{\upshape{(Probability Distribution over SLD Derivations)}}
\label{def:distrSLD}
Let $d = (G_0, G_1, \dots, G_n)$ be a finite SLD derivation. The probability of $d$ is defined as:
\[
p(d) = \prod_{i=0}^{n-1} p(G_{i+1} \mid G_i)
\]
where $p(G_{i+1} \mid G_i)$ is the probability of transitioning from goal $G_i$ to goal $G_{i+1}$ by applying one SLD resolution step.
We define $p(G_{i+1} \mid G_i)$ as the \emph{resolution transition model}, which specifies a distribution over the possible outcomes of resolving the leftmost atom in $G_i$.
\end{definition}

Given a query $q$, we are interested in computing its success probability under the derivation probability distribution. 

\begin{definition}[Success Probability of a Query]
\label{def:successSLD}
Let $\mathcal{R}(q)$ be the set of all successful derivations for a query $q$, i.e., derivations $d = (G_0, G_1, ..., G_n)$ with $G_0=q$ and $G_n=\textit{True}$, for some $n>0$.
The success probability of $q$ is:
\[
p_{success}(q) = \sum_{d \in \mathcal{R}(q)}  p(d) = \sum_{d \in \mathcal{R}(q)} \prod_{i=0}^{n-1} p(G_{i+1} \mid G_i) \ .
\]
\end{definition}

\textbf{Stochastic Logic Programs. }Pure normalized SLPs extend logic programming by associating each clause \( C \) with a fixed probability \( \lambda_C \in [0, 1] \), written as \( \lambda_C :: H \leftarrow B_1, \ldots, B_m \). For a predicate \( r \), let \( \mathcal{C}(r) \) denote the set of clauses with head predicate \( r \), satisfying \( \sum_{C \in \mathcal{C}(r)} \lambda_C = 1 \). This defines a \emph{scalar parameterization} of the resolution transition model: \( p(G_{i+1} \mid G_i) = \lambda_{C_i} \), where \( C_i \) is the clause used in the resolution step \( G_{i+1} = \text{res}(G_i, C_i, \theta_i) \).

\textbf{DeepStochLog.}
DeepStochLog replaces fixed clause probabilities of SLPs with a neural network $\rho_\lambda$, which produces a probability distribution over possible variable substitutions. 
This induces a \emph{term-level neural parameterization} of the transition model
$p(G_{i+1} \!\mid \!G_i) \!=\! \rho_\lambda(X\theta_i,Y\theta_i)$,
where $\theta_i$ is the MGU used in the resolution step from $G_i$ to $G_{i+1}$, i.e., $\rho_\lambda$ is conditioned on the instantiated input and output terms $X\theta_i,Y\theta_i$ realizing the unification step. An example is shown in Figure \ref{fig:param_compare}.

\textbf{Limitations.}
SLPs and DeepStochLog both restrict the expressiveness of stochastic SLD resolution. In classical SLPs, clause probabilities $\lambda_C$ are fixed per predicate and do not depend on the specific substitutions from unification, making resolution insensitive to query context or input variation. DeepStochLog extends this by conditioning clause probabilities on subsymbolic inputs (e.g., differing scores for \texttt{country(img32, Y)} vs. \texttt{country(img73, Y)}), but it requires a fixed output domain, limiting flexibility when valid substitutions depend on the query (e.g., \texttt{neighborOf(italy, Y)} vs. \texttt{neighborOf(japan, Y)}). Additionally, both methods restrict clause selection to the leftmost atom, preventing the use of global context for more informed decisions.

\subsection{Markov Decision Processes}
An MDP is a formal model of decision-making defined by the tuple $(\mathcal{S}, \mathcal{A}, P, R, \gamma)$, where $\mathcal{S}$ is a set of states, $\mathcal{A}(s)$ is a set of actions available in state $s$, $P(s,a,s')$ is the transition function giving the probability of transitioning from state $s$ to $s'$ with action $a$, $R(s, a, s')$ is the reward function giving the expected reward for the transition, and $\gamma \in [0, 1]$ is the discount factor for future rewards. A policy $\pi(a|s)$ defines the probability of selecting action $a$ in state $s$. Given a starting state $s_0$, solving an MDP requires finding the optimal policy $\pi^*$ that maximizes the future cumulative reward, defined as the value function: 
\[
v(s_0) = \mathbb{E} \left[ \sum_{t=0}^{\infty} \gamma^t R_{t+1} \Big| S=s_0 \right]
\]
where the expectation is over trajectories of states and actions sampled from $P(s,a,s')$ and $\pi(a|s)$.

\section{\our{}} \label{sec:method}
To address the limitations of previous approaches, we propose a shift in perspective: instead of parameterizing the logic program itself (e.g., via clause weights), we parameterize the resolution process. To this end, we introduce \our{}, which defines clause selection probabilities conditioned on the entire current goal and enables dynamic adaptation to varying substitution spaces. \our{} is a framework for differentiable, goal-conditioned SLD resolution that supports flexible and context-aware reasoning.

We define the resolution transition model in \our{} as:
\begin{equation}
\label{eq:resTransOur}
    p_\lambda(G_{i+1} \mid G_{i}) = \frac{\exp(f_\lambda(G_{i}, G_{i+1}))}{\displaystyle\sum_{G \in \mathcal{N}(G_{i})} \exp(f_\lambda(G_{i}, G))}
\end{equation}
where $\mc{N}(G)$ denotes the set of next goals from resolving the leftmost atom of $G$; $f_\lambda(G, G^\prime)$ scores the compatibility between the learned embeddings of the current goal $G$ and the candidate next goal $G^\prime$ (as defined in the next paragraph); and $\lambda$ comprises all model parameters.

\paragraph{Goal Representation.}
To enable neural processing of symbolic goals, we construct goal embeddings hierarchically from (sub)symbolic and atomic components.

\textit{(Sub)symbolic Embeddings. } 
Let $\mathcal{V}$ denote the set of all entities involved in reasoning, including symbolic tokens (e.g., predicates, constants, variables) and subsymbolic inputs (e.g., images).
Each element $v \in \mathcal{V}$ is associated with a learnable dense embedding $\mathbf{e}_v \in \mathbb{R}^d$.  
The embeddings $e_v$ for all $v \in \mathcal{V}$  are part of $\lambda$ (cf. Equation~\ref{eq:resTransOur}) and are trained jointly with the reasoning model.
This design enables a unified embedding space in which both symbolic and subsymbolic information can interact seamlessly during inference and learning.

\textit{Atom Embeddings. }
For an atom $A = r(t_1, \ldots, t_n)$ with predicate $r$ and arguments $t_1, \ldots, t_n$, we compute its embedding via a composition function:
$
\mathbf{e}_A = f_{\text{atom}}(\mathbf{e}_r, \mathbf{e}_{t_1}, \ldots, \mathbf{e}_{t_n})
$,
where $f_{\text{atom}}$ is a learnable and differentiable function that maps the component embeddings into an atom embedding. Since the arity of a predicate is fixed, \( f_{\text{atom}} \) can be instantiated with any neural network or knowledge graph embedding model.

\textit{Goal Embeddings. }
The embedding of a goal $G = {} \leftarrow A_1, \ldots, A_n$ is computed by aggregating the embeddings of its constituent atoms:
$
\mathbf{e}_G = f_{\text{agg}}(\mathbf{e}_{A_1}, \ldots, \mathbf{e}_{A_n})
$,
where $f_{\text{agg}}$ is a differentiable aggregation function. Common choices include:
sum, mean, or a learnable neural aggregation.

\textit{Compatibility Score. }
The compatibility score between goals $G$ and $G^\prime$ is computed as the scalar product of their embeddings:
$
f_\lambda(G, G^\prime) = \mathbf{e}_G \cdot \mathbf{e}_{G^\prime} 
$.

\paragraph{Learning.}
In line with similar approaches (cf. Section~\ref{sec:prel}), \our{} defines the query success probability as in Stochastic SLD resolution (cf. Definition~\ref{def:successSLD}). Moreover, our learning objective aligns with DeepStochLog, 

\begin{equation}
\label{eq:objective}
\mathcal{L}(\lambda) = \sum_{(q^{(i)}, y^{(i)}) \in \mathcal{D}} \ell\left(p_{\text{success}}^{\lambda}(q^{(i)}), y^{(i)}\right)
\end{equation}
where \(\ell(p, y) = (1 - 2y) \cdot p\) is a linear loss prompting high success probability for positive queries (\(y=1\)) and low for negative ones (\(y=0\)); $p_{\text{success}}^{\lambda}(q^{(i)})$ denotes the success probability of query $q^{(i)}$ under the \our{} parameterization; and \(\mathcal{D} = \{(q^{(i)}, y^{(i)})\}_{i=1}^{N}\) is the labeled dataset.

\paragraph{Expressiveness.}
Before discussing learning and inference capabilities, we show in the following theorem that \our{} can approximate any probability distribution over SLD derivations arbitrarily well. 

\begin{proposition}[Universal Approximation]
\label{the:approx}
    Any probability distribution over SLD Derivations can be approximated with arbitrary precision by \our{}.
\end{proposition}
The proof is in Appendix A.1.

\section{Markov Decision Processes for \our{}}
\label{sec:mdpmodelling}
\our's novel perspective on parameterizing stochastic SLD resolution opens the door to interpreting learning and reasoning in any (deep) SLP as learning an optimal policy within a logic program-based environment of an MDP. In the following we show how \our{} instantiates an MDP by defining the components $(\mathcal{S},\mathcal{A},P,R,\gamma)$.

\paragraph{SLD-resolution as MDP.}
We formalize the SLD-resolution as an MDP,
where each component is defined based on a dataset of labeled queries \(\mathcal{D} = \{(q^{(i)}, y^{(i)})\}_{i=1}^{N}\), with \(y^{(i)} \in \{1, 0\}\) indicating whether \(q^{(i)}\) is a positive or negative query.

\textbf{States. }
The state space \(\mathcal{S}\) is defined by $\mathcal{S} = \mathcal{Q} \cup \mathcal{G} \cup \{\textit{True},\textit{False}\}$, 
where
  \(\mathcal{Q} = \{q^{(i)}\}_{i=1}^{N}\) is the set of initial queries in $\mc{D}$,
  and \(\mathcal{G}\) is the set of all intermediate sub-goals that can arise from any derivation starting from any \(q^{(i)}\).

\textbf{Actions. }
Let $G={}
\leftarrow A_1, \dots, A_n$ be a state, and let $\mc{P}$ be the definite logic program. The action space \(\mathcal{A}(G)\) is defined as $\mathcal{A}(G) = \left\{(C, \theta)\middle|\; C \in \mathcal{P},\theta = \text{MGU}(A_1, \text{head}(C)) \,\right\}$. Additionally, we introduce the special $\textit{False}$ action to enable the model to learn early abandonment. More details on the advantages and implementation
are provided in Appendix B.2.

\textbf{Transition Function. }
The transition function is deterministic and follows from logic resolution: given $G_{i+1}=\operatorname{res}(G_i, C, \theta)$, the transition probability is defined as
$P(G' \mid G_i, C, \theta) = \mathbf{1}\{G'=G_{i+1}\}.$

\textbf{Reward. }
To define the reward, we introduce an auxiliary set of labeled states, where each state is paired with the label of the original query it stems from, i.e., $\tilde{\mathcal{S}} = \{(G,y^{(i)}) \mid G \in \mathcal{S},y^{(i)} \in \{0,1\} \text{ is the label of } q^{(i)} \in \mathcal{Q} \}$. Note that the labeled states \(\tilde{\mathcal{S}}\) are only used for reward computation, do not affect the state space \(\mathcal{S}\), and are not accessed by the policy.
The reward function \(R : \tilde{\mathcal{S}} \to \mathbb{R}\) is defined as
\begin{equation}
\label{eq:reward}
R(G, y^{(i)}) = 
\begin{cases}
2 y^{(i)} - 1 & \text{if } G = \textit{True} \\
0 & \text{otherwise}
\end{cases}    
\end{equation}
Hence, proving a positive query yields a reward of $+1$, while a negative query yields $-1$. No immediate reward is given for intermediate states.

\textbf{Policy and Value Function. }
In our MDP formulation, the policy \(\pi_\lambda\) defines a probability distribution over $\mc{A}(G_i)$ for a goal $G_i$. Each action $a=(C, \theta) \in \mc{A}(G_i)$ to $G_i$ deterministically yields the next goal $G_{i+1}=\operatorname{res}(G_i, a)$,
establishing a one-to-one mapping between actions and successor states. Formally, this yields: 
\[
\pi_\lambda(a \mid G_i) = p_\lambda(G_{i+1} \mid G_i)
\]
This equation highlights that a policy guiding the SLD resolution process is parameterized exactly by the transition probabilities of our resolution model (cf. Equation~\ref{eq:resTransOur}). This correspondence arises from the semantics of SLPs: clause probabilities represent the likelihood that a clause is selected by a prover - here modeled as an agent governed by \(\pi_\lambda\) - when proving a query.

As in standard MDPs, the value function of a policy is defined as its expected cumulative reward. While value functions generally follow the Bellman recursion, our SLD resolution MDP allows a simplified formulation for the value of a non-terminal goal $G$ with label $y$ (see Appendix A.2 for the derivation):
\begin{eqnarray}
\label{eq:bellman}
V^{\pi_\lambda}(G, y)
= & \displaystyle \sum_{a \in \mathcal{A}(G)} \pi_\lambda(\operatorname{res}(G, a), G) \cdot\\
& ~~~~~~~~~~~\cdot ~ V^{\pi_\lambda}(\operatorname{res}(G, a), y) \ , \nonumber
\end{eqnarray}
with boundary condition \(V^{\pi_\lambda}(G, y) = R(G, y)\) for a terminal goal \(G\).

\paragraph{Example. } We illustrate the introduced terms with the following example. Consider a logic program \( \mathcal{P} = \{C_1, C_2, C_3, C_4, C_5\} \), where the clauses are defined as:  
$C_1 = \textit{locIn}(X, Z) \leftarrow \textit{neighOf}(X, Y), \textit{locIn}(Y, Z)$, $C_2 = {} \leftarrow \textit{neighOf}(\textit{it}, \textit{fr})$,  
$C_3 = {}\leftarrow \textit{locIn}(\textit{fr}, \textit{eu})$, $C_4 = {}\leftarrow \textit{locIn}(\textit{tr}, \textit{gr})$,  
and $C_5 = {}\leftarrow \textit{locIn}(\textit{gr}, \textit{eu})$. Consider a positive query \( q^{(0)} = G_0^{(0)} = {}\leftarrow \textit{locIn}(\textit{it}, \textit{eu}) \) with label \( y^{(0)} = 1 \). The corresponding action space is \( \mathcal{A}(G_0^{(0)}) = \{(C_1, \{X/\textit{it}, Z/\textit{eu}\})\} \). Applying the action gives the next goal \( G_1^{(0)} = {}\leftarrow \textit{neighOf}(\textit{it}, Y), \textit{locIn}(Y, \textit{eu}) \), with intermediate reward \( R(G_1^{(0)}) = 0 \). Assuming the episode ends after \( n \) resolution steps with \( G_n^{(0)} = \textit{True} \), the final reward is \( R(G_n^{(0)}, y^{(0)}) = 2 \cdot 1 - 1 = 1 \). Now consider a negative query \( q^{(1)} = G_0^{(1)} = {}\leftarrow \textit{locIn}(\textit{tr}, \textit{eu}) \) with label \( y^{(1)} = 0 \). If the episode ends in \( G_m^{(1)} = \textit{True} \) after \( m \) resolution steps, the final reward is \( R(G_m^{(1)}, y^{(1)}) = 2 \cdot 0 - 1 = -1 \).

\section{Learning in \our{}}
\label{sec:learning}

Learning in \our{} is equivalent to learning the optimal policy of the corresponding MDP defined in Section~\ref{sec:mdpmodelling}. For the empirical distribution defined by $\mathcal{D}$, the objective is defined as
$
J(\pi_{\lambda}) = \sum_{i=1}^{N}[V^{\pi_\lambda}(q^{(i)}, y^{(i)})] 
$ \cite{sutton1999reinforcement}.
% where \(\mu(q)\) is a uniform distribution over the set of initial queries \(\mathcal{Q} = \{q^{(i)}\}_{i=1}^{N}\), and \(V^{\pi_\lambda}\) is the value function under policy \(\pi_\lambda\). 
Given our reward design in Equation~\ref{eq:reward} and setting the discount factor \(\gamma = 1\), we show in Appendix A.3 that the objective function simplifies to:
\[
J(\pi_\lambda) =  \sum_{i=1}^{N} (2y^{(i)} - 1) \cdot p_{\text{success}}^{\pi_\lambda}(q^{(i)}) \ 
\]
where $p_{\text{success}}^{\pi_\lambda}(q^{(i)})$ is the success probability of $q^{(i)}$ under $\pi_\lambda$.

\begin{proposition}
Maximizing the expected return $J(\pi_\lambda)$ in our MDP formulation corresponds to minimizing the objective $\mc{L}(\lambda)$ defined in Equation~\ref{eq:objective}.
\end{proposition}
Maximizing $J(\pi_\lambda)$ also aligns with minimizing the cross-entropy loss commonly used in NeSy learning; see Appendix A.3 for the proof.

\textbf{Exact Inference via Dynamic Programming. } As demonstrated in our experiments, standard NeSy benchmarks, often perceived to involve combinatorial search spaces, can be encoded to yield a manageable state space. In such cases, dynamic programming enables significantly better scalability without introducing approximation. By reasoning in terms of the state-space size of the corresponding MDP, our formulation helps to design nested symbolic components that keep the size tractable and support efficient exact inference.

\textbf{Approximate Inference via Policy Gradient. }
When the full goal space is intractable, the mapping between SLP derivations and MDPs enables the application of reinforcement learning techniques to learn the clause selection policy. Prominent algorithms such as REINFORCE \cite{williams1992simple}, Actor-Critic \cite{sutton2000policy}, and Proximal Policy Optimization (PPO) \cite{schulman2017proximal} can be adapted. For example, we employ PPO to optimize the neural policy, using the standard clipped surrogate objective. Alongside the policy, we parameterize a neural value function with the same architecture to estimate returns and reduce variance during training. While value-based methods are also applicable in principle, we leave their exploration for future work.

\section{Experiments} \label{sec:exp}
We present two sets of experiments. First, we assess our approach on a neurosymbolic task that learns perception given the background knowledge, showing that \our{} (\acr{}) outperforms advanced approximate inference systems in both accuracy and scalability while scaling better than exact inference systems such as DeepStochLog. The second set of experiments addresses explainable knowledge graph (KG) completion. Here, \acr{} learns from data to generate high-probability proofs for positive queries and low-probability proofs for negative ones. The fact that each classification is expressed by a proof makes \acr{}'s decisions fully interpretable. Comparative experiments against state-of-the-art proof-based KG systems show that our method scales to large KGs without compromising its predictive performance. All code and data will be publicly available upon publication. 

\subsection{MNIST Addition} \label{exp:mnist}
\begin{table*}[ht!]
\centering
{
\begin{tabular}{@{}lllccccc@{}}
& N && 4 & 15 & 100 & 200 & 500 \\
\midrule
&Reference  && 92.9 & 75.8 & 15.4 & 2.4 & 0.009\\
\midrule
\multirow{7}{*}{\rotatebox{90}{NeSy Exact}}&KLay  &Time&  T/O \\\cline{2-8}
&\multirow{2}{*}{DeepStochLog} & Acc. & $92.7\pm0.6$ & \multirow{2}{*}{T/O}  \\ 
&& Time & $141.2\pm30.8$ \\
\cline{2-8}
&\multirow{2}{*}{DeepSoftLog}  & Acc. & $93.5\pm0.6$ & $77.1\pm1.6$ & \multirow{2}{*}{T/O} \\
&& Time &$879.9\pm45.3$&$1324.9\pm72.6$\\ \cline{2-8}
&\multirow{2}{*}{\acr{}~(DP)} & Acc.& $94.0\pm0.3$ & $80.8\pm1.1$ & $37.8\pm2.9$ & $23.2\pm3.0$ & $6.0\pm4.9$\\
&& Time & $25.4\pm6.3$ & $41.8\pm3.7$ &$105.8\pm21.8$&$128.0\pm33.3$&$469.0\pm51.6$\\ 
\midrule
\multirow{8}{*}{\rotatebox{90}{NeSy Approximate}}&\multirow{2}{*}{Scallop} & Acc. & $92.3\pm0.7$ & \multirow{2}{*}{T/O} \\
&& Time & $73.6\pm3.9$ &&&&\\ \cline{2-8}
&\multirow{2}{*}{A-NeSI}  & Acc. & $92.6\pm0.8$ &$75.9\pm2.2$& \multirow{2}{*}{T/O} \\
&& Time & $70.1\pm3.1$&$916.5\pm35.8$\\ \cline{2-8}
&\multirow{2}{*}{EXAL}  & Acc. & $91.8\pm0.8$ &$73.3\pm2.1$& \multirow{2}{*}{T/O} \\
&& Time &$32.0\pm1.7$&$145.1\pm6.8$&&&\\
\cline{2-8}
&\multirow{2}{*}{\acr{}~(PG)} & Acc. & $93.5\pm0.4$ & $76.9\pm1.9$ & $0.0\pm0.0$ & $0.0\pm0.0$ & $0.0\pm0.0$  \\
&& Time & $25.7\pm8.6$&$49.7\pm12.0$&$282.6\pm20.5$&$295.8\pm46.0$&$865.0\pm67.9$\\ 
\bottomrule
\end{tabular}
}  % end \small
\caption{Test accuracy and training times (in minutes, shown in parentheses) on MNIST addition with sequence length \(N\). A time-out (T/O) of 24 hours (1440 minutes) is applied. Reference indicates the accuracy on the sum obtainable by a single-digit classifier with accuracy of $0.9907$, e.g., Reference=$0.9907^{2N}$~\cite{maene2023soft}.
}
\label{tab:MNISTaddition}
\end{table*}

\begin{table*}[ht]
\centering
{
\begin{tabular}{l l c c c c | c c c c}
& & \multicolumn{4}{c}{Family} & \multicolumn{4}{c}{WN18RR} \\ \cmidrule{3-10}
&        & MRR & Hits@1 & Hits@3 & Hits@10 & MRR & Hits@1 & Hits@3 & Hits@10 \\ \midrule
\multirow{1}{*}{\rotatebox{90}{KGE}} 
& ComplEx & 0.964 & 0.938 & 0.991 & 0.994 & 0.479 & 0.438 & 0.478 & 0.540 \\  
& RotatE & 0.968 & 0.943 & 0.994 & 0.995 & 0.833 & 0.787 & 0.863 & 0.913 \\ \midrule
\multirow{4}{*}{\rotatebox{90}{NeSy}}
& R2N$_{SG}$ & 0.985 & 0.981 & 0.989 & 0.989 & 0.724 & 0.699 & 0.733 & 0.755 \\
& DCR$_{SG}$ & 0.975 & 0.956 & 0.994 & 0.995 & 0.817 & 0.754 & 0.862 & 0.922 \\
& SBR$_{SG}$ & 0.981 & 0.966 & 0.995 & 0.995 & 0.840 & 0.784 & 0.879 & 0.935 \\
& \acr{}~(PG) & 0.986 & 0.979 & 0.994 & 0.995 & 0.834 & 0.784 & 0.868 & 0.918 \\
\midrule
\end{tabular}
\caption{Comparison of KGE and NeSy Methods on Family and WN18RR Datasets.}
}
\label{tab:kg_comparison}
\end{table*}

MNIST addition is a popular NeSy benchmark, where the model predicts the sum of two numbers represented as sequences of $N$ MNIST digit images. The only supervision provided is the final sum, without explicit labels for individual digits. This task serves as a testbed for evaluating the scalability of \acr{}, as the reasoning complexity increases with the number of digits $N$. %We follow the experimental setup as in \cite{manhaeve2018deepproblog}.

We explore both dynamic programming (DP) and policy gradient (PG) approaches for this task. For DP, we compute the exact optimal policy by maximizing the value function of the MDP induced by the logic program, following the formulation in Equation~\ref{eq:bellman}.
For PG, we implement an off-policy REINFORCE algorithm where the policy is a neural digit classifier. The logic program structure naturally constrains the set of valid actions at each step, reducing sampling variance. The logic programs are provided in Appendix C.1.

We compare with state-of-the-art NeSy methods on predicted sum accuracy and training time. Baselines include three exact systems: KLay \cite{DBLP:conf/iclr/MaeneDM25}, DeepStochLog, and DeepSoftLog \cite{maene2023soft}; and three approximate systems: Scallop \cite{huang2021scallop}, A-NeSI \cite{van2023nesi}, and EXAL \cite{verreet2024explain}. Results are summarized in Table~\ref{tab:MNISTaddition}, with experimental setups, hyperparameters, and additional intermediate N values provided in Appendix C.1.

The DP variant achieves accuracy comparable to existing exact NeSy systems while significantly outperforming both exact and approximate baselines in terms of training efficiency. We observed that no other previous method successfully scales to $N = 100$ without timing out. In the approximate setting, our PG variant achieves the highest accuracy. It scales beyond $N = 100$, demonstrating strong computational efficiency, but converges to a suboptimal policy due to the vast combinatorial search space.

\subsection{Explainable Knowledge Graph Completion}

\begin{figure}[th!]
\centering

% --- Top Proof (WN18RR) ---
% This tree is wide, so we resize it to the full column width.
% This causes the font to shrink significantly.
\begin{forest}
  for tree={
    l sep=4mm,
    s sep=1mm,
    align=center,
    font=\small
  }
  [{synset\_domain\_topic\_of\\(09788237, 08441203)}
    [{hypernym\\(09788237, 09774266)}]
    [{synset\_domain\_topic\_of\\(09774266, 08441203)}
      % second row
      [{synset\_domain\_topic\_of\\(00872886, 08441203)}
        % third row
        [{derivationally\_rel\_form\\(00872886, 09774266)}]
        [{synset\_domain\_topic\_of\\(09774266, 08441203) $\square$}]
      ]
      % the other child
      [{derivationally\_rel\_form\\(09774266, 00872886)}]
    ]
  ]
\end{forest}%

\vspace{1em}
\centerline{\rule{0.8\columnwidth}{0.4pt}}
\vspace{1em}

% --- Bottom Proof (Family) ---
% CHANGE: We resize this to 0.42\columnwidth instead of full width.
% Since this tree is naturally much narrower (about 40% of the top tree's width),
% resizing it to 0.42\columnwidth applies roughly the same scaling factor
% as the top tree, ensuring the text sizes look identical.

\begin{forest}
  for tree={
    l sep=4mm,
    s sep=1mm,
    align=center,
    font=\small
  }
  [{uncle(2266, 2252)}
    [{brother(2260, 2252)}]
    [{uncle(2266, 2260)}
      [{sister(2262, 2260)}]
      [{uncle(2266, 2262) $\square$}]
    ]
  ]
\end{forest}%

\caption{Examples of multi-step proof trees discovered by \our{} for the Family and WN18RR datasets.}
\label{tab:proof_tree_example}
\end{figure}
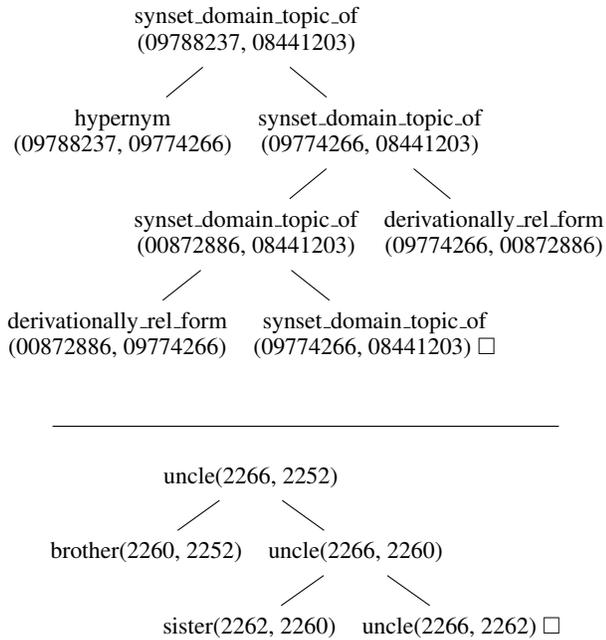

We evaluate DPrL on popular KG completion benchmarks: Family \citep{kok2007statistical} and WN18RR \citep{dettmers2018convolutional}. Family models familial relationships, while WN18RR is a challenging subset of WordNet with inverse relations removed (dataset statistics in Appendix C.2). All methods were tested with 200 randomly sampled negative corruptions per query.

Table 2 compares \acr{} to KGE baselines (ComplEx \cite{trouillon2016complex}, RotatE \cite{sun2019rotate}) and logic-based NeSy systems: SBR \cite{diligenti2017semantic}, R2N \cite{marra2025relational_reasoning_networks} and DCR \cite{barbiero2023interpretable}.
Please note that these NeSy systems would not scale to the considered datasets without relying on custom approximate grounding techniques \cite{groundingmethods}
%, which make it possible to perform approximate reasoning using these methodologies.
We indicate these variants as $\textit{R2N}_{\text{SG}}$, $\textit{DCR}_{\text{SG}}$, and $\textit{SBR}_{\text{SG}}$. 
However, these fast grounding techniques require manual parameter tuning, whereas \acr{} automatically learns the best exploration strategy for each dataset.

\acr{} uses Proximal Policy Optimization (PPO) \citep{schulman2017proximal} to navigate the complex, large-scale action spaces emerging in KG reasoning. The model computes the scores $p^\lambda_{\text{success}}(q)$ for each query $q$ so that the standard evaluation metrics for each task (MRR, Hits@N) can be computed (see Appendix C.2 for the metric definitions).
All NeSy methods, including \acr{}, use RotatE output as a prior, learning a logic-based adjustment on top of it.

On Family, all NeSy systems improve performance over the baseline. \acr{} is particularly effective, especially for MRR and Hits@1. A similar trend is observed on WN18RR, where \acr{} and SBR outperform other methods. 

Unlike the partially latent reasoning done by R2N or DCR, \acr{} is fully interpretable. %, as each classification decision is the result of the execution of a proof that can be inspected by a human. 
% Table~\ref{tab:proof_tree_example} showcases two examples of explanations in the form of proofs discovered by \acr{}: a depth-3 proof for the query synset\_domain\_topic\_of(09788237, 08441203) from the WN18RR dataset, and a proof for the query uncle(2266, 2252) in the Family dataset.
Figure~\ref{tab:proof_tree_example} presents two proof trees generated by \acr{}: a depth-3 proof for synset\_domain\_topic\_of(09788237, 08441203) from WN18RR, and a proof for uncle(2266, 2252) from Family.

Methods based on a complete or approximate grounding technique, like the ones proposed by \cite{groundingmethods}, would need to instantiate and process a combinatorially large number of ground formulas to find a proof at a high depth. In contrast, \acr{} can avoid an exhaustive search by learning an efficient proof-finding policy, which navigates deep reasoning paths efficiently. %For example, on the Family dataset, the smart grounding techniques proposed by \cite{groundingmethods} instantiate $18071$, $45466$, and $181095$ groundings for the test queries when allowing proofs at depths 1, 2, and 3, respectively. On the other hand, \acr{} needs to evaluate only $12334$ ground formulas to perform full inference over the test set.
For instance, on the test set of Family, the smart grounding methods of \cite{groundingmethods} instantiate 18,071, 45,466, and 181,095 groundings at depths 1, 2, and 3, respectively. In contrast, DPrL needs just 12,334 ground evaluations.

% In conclusion, \acr{} demonstrates superior scalability over other NeSy models capable of learning perception and structure, such as DeepSoftLog (which fails to scale to the datasets presented here), while preserving the explainability and interpretability advantages of logic-based NeSy approaches.

In conclusion, \acr{} demonstrates superior scalability over other perception- and structure-learning NeSy models, such as DeepSoftLog (which fails to scale to the datasets presented here), while preserving the explainability and interpretability advantages of logic-based NeSy approaches.

\section{Related Work}
\label{sec:rel_work}

\paragraph{DeepStochLog.}

DeepStochLog extends SLPs by parameterizing clause probabilities via neural networks under derivation-based semantics. However, its exact inference leads to the combinatorial explosion of proof paths, making it intractable for complex tasks. It also assumes fixed output domains and conditions neural predictions on partial goals, limiting flexibility and context awareness. In contrast, \acr{} factorizes the derivation distribution over resolution steps, conditions each decision on the full current goal, and supports variable output domains. This results in better scalability (linear in derivation steps and matching unification), more informed decisions, and greater expressivity.

\paragraph{Scaling inference in NeSy.}
Efforts to improve the scalability of exact NeSy systems include DeepStochLog, which adopts a more efficient derivation-based semantics, and KLay \cite{DBLP:conf/iclr/MaeneDM25}, which introduces GPU-parallel arithmetic circuits.
Approximate NeSy methods include search-based techniques \cite{manhaeve2021approximate, huang2021scallop} and random walks with restart \cite{wang2013programming}, which limit inference to parts of the proof space but risk bias. Sampling-based approaches \cite{verreet2023explain} avoid full model counting but are still limited by grounding. Variational inference \cite{abboud2020learning, dos2021neural, van2023nesi} replaces logic inference with learned predictors, trading interpretability for speed. Parametrized grounding \cite{castellano2025grounding} offers a tunable balance between scalability and expressivity but lacks general guarantees provided by \acr{}.

\paragraph{Automated Theorem Proving.}
Automated Theorem Proving (ATP) finds a proof for a query within deterministic logical frameworks. Recent advances integrate learning-based heuristics to guide proof search \cite{rocktaschel2017end,kaliszyk2018reinforcement,rawson2019neurally,zombori2020prolog,lample2022hypertree}. Unlike ATP systems, \acr{} operates in a probabilistic setting and learns from labeled data. Rather than determining provability, it optimizes inference by maximizing the probability of positive queries and minimizing that of negatives.

\paragraph{Reinforcement Learning in KG Completion.}
Reinforcement learning (RL) has been applied to knowledge graph (KG) completion mainly via path-finding models like DeepPath \cite{xiong2017deeppath} and MINERVA \cite{das2017go}, which train agents to navigate KGs by sequentially selecting entities and relations. These approaches map relation-path discovery to an MDP but are specific to KG structures. While paths can serve as explanations, they may be complex and hard to interpret. In contrast, \our{} maps SLPs to MDPs, enabling RL-guided proof search across any domain expressible with SLPs, offering greater generality beyond KGs.

\section{Conclusions}
\label{sec:conc}
We introduce \our{} (\acr), a scalable and flexible NeSy framework that bridges expressive logical representations and efficient data-driven learning. \acr{} leverages SLPs with derivation-based semantics and introduces a more expressive neural parameterization by conditioning on the full goal at each resolution step. A primary contribution is the novel and formal correspondence between inference and learning in our deep SLPs and policy optimization in MDPs, enabling dynamic programming and RL to tackle previously intractable NeSy problems. 

The presented method shares a limitation with prior work: clause selection is restricted to the leftmost atom. While this preserves expressivity and completeness of the proofs, we plan to explore a more flexible selection, which could more efficiently prune failing derivations. Our current implementation uses standard policy gradients; future work will consider advanced RL techniques and broader evaluation settings. We also see potential in extending to possible world–based NeSy methods like Markov Logic Networks \cite{richardson2006markov} and in studying automatic logic program reformulations to reduce the search space \cite{leuschel2002logic}.

\section*{Acknowledgments}
This work has been supported by the EU Framework Program for Research and Innovation Horizon under the Grant Agreement No 101073307 (MSCA-DN LeMuR), the European Research Council (ERC) under the European Union’s Horizon 2020 research and innovation program (Grant agreement No. 101142702), and the Flemish Government under the “Onderzoeksprogramma Artificiële Intelligentie (AI) Vlaanderen” programme.
F. Giannini acknowledges support from the Partnership Extended PE00000013 - FAIR - Future Artificial
Intelligence Research - Spoke 1 Human-centered AI and ERC-2018-ADG G.A. 834756 XAI: Science
and technology for the eXplanation of AI decision making. 
L. De Raedt is also supported by the Wallenberg AI, Autonomous Systems and Software Program (WASP) funded by the Knut and Alice Wallenberg Foundation.
M. Diligenti and M. Gori are also supported by the project ``CONSTR: a COllectionless-based Neuro-Symbolic Theory for learning and Reasoning'', PARTENARIATO ESTESO ``Future Artificial Intelligence Research - FAIR'', SPOKE 1 ``Human-Centered AI'' Universit\`a di Pisa,  ``NextGenerationEU'', CUP I53C22001380006.
G. Marra acknowledges support by FWO (G033625N) and KU Leuven Research Fund (CELSA, CELSA/24/008).

Y. Jiao would like to thank Gabriele Venturato and Lennert De Smet for stimulating discussions.

\bibliography{references}

\newpage

\setcounter{secnumdepth}{2}
\renewcommand\thesubsection{\thesection.\arabic{subsection}}
\renewcommand\labelenumi{\thesubsection.\arabic{enumi}}

\appendix

\section{Proofs}

\subsection{Proof of Proposition 1}
\label{app:proof}

%\begin{proof}
\begin{definition}[Derivation Tree]
Let $\mathcal{T} = (\mathcal{V}, \mathcal{E})$ be a tree of derivations rooted at $G_0 \in \mathcal{V}$, where each node $G \in \mathcal{V}$ is a goal with a unique representation $\mathbf{e}_G$. We assume all nodes are unique (no goal appears twice in the derivation).
\end{definition}

\begin{definition}[Target and network distributions]
Let $p(G^\prime | G)$ be the distribution computed by the network for 
% $\forall G \in \mathcal{V} \land G^\prime \in \mathcal{A}(G)$ 
$\forall (G, G') \in \mathcal{E}$, and $p^t(G^\prime | G) \in (0,1]$ be the target distribution we want to approximate.
\end{definition}

We aim to show that for any valid target probability distribution $p^t$ and any $\epsilon > 0$, there exist network parameters such that:
\[
|p(G^\prime | G)- p^t(G^\prime | G)| < \epsilon \ \quad \forall (G, G') \in \mathcal{E} .
\]

\begin{proof}
Given the definition of $p(G^\prime|G)$ in Section~\ref{sec:method}, we need to find scalar products $\mathbf{e}_G \cdot \mathbf{e}_{G^\prime}$ such that:
\[
p(G^\prime|G) = \frac{ \exp(\mathbf{e}_{G^\prime} \cdot \mathbf{e}_G) } {\displaystyle\sum_{G_c \in \mathcal{A}(G)} \exp(\mathbf{e}_{G_c} \cdot \mathbf{e}_{G}) } \approx p^t(G^\prime | G) \ .
\]
If $\exp(\mathbf{e}_{G^\prime} \cdot \mathbf{e}_G) \approx p^t(G^\prime | G), ~\forall (G, G') \in \mathcal{E}$, then the model is a good approximation of the target:
\begin{eqnarray*}
p(G^\prime|G) &=& \frac{ \exp(\mathbf{e}_{G^\prime} \cdot \mathbf{e}_G) } {\displaystyle\sum_{G_c \in \mathcal{A}(G)} \exp(\mathbf{e}_{G_c} \cdot \mathbf{e}_{G}) }  \\
&\approx& \frac{ \exp(\mathbf{e}_{G^\prime} \cdot \mathbf{e}_G) } {\displaystyle\sum_{G_c \in \mathcal{A}(G)} p^t(G_c | G) } = \\
& = & \exp(\mathbf{e}_{G^\prime} \cdot \mathbf{e}_G) \approx p^t(G^\prime | G) \ .
\end{eqnarray*}
Therefore, $\mathbf{e}_{G^\prime} \cdot \mathbf{e}_G = \log p^t(G^\prime | G)$ are sufficient conditions to find a perfect approximation for the target distribution.

We now show that it is possible to construct the embeddings $\mathbf{e}_G$ so that they respect such constraints. Let $d$ be an embedding size such that $b_G \in \mathbb{R}^d$ is an orthonormal basis for the goals $G$.
This basis exists by assuming that the embedding space is at least as large as the number of nodes (distinct derivation steps) in the tree, e.g., $d \ge |\mathcal{V}|$.

If this holds, the goal representations can be recursively constructed by first defining the embedding for the root node $G_0$ to be its own basis vector: $\mathbf{e}_{G_0} = b_{G_0}$.
Then, we define the embeddings to be recursively computed as:
$\mathbf{e}_{G^\prime} = \log p^t(G^\prime | G) \cdot b_G + b_{G^\prime}$.
Since the nodes $G, G^\prime$ are distinct (and so is also $G$'s parent), their corresponding basis vectors are orthogonal, and it can be shown by basic math that:
\begin{align*}
    \mathbf{e}_{G^\prime} \cdot \mathbf{e}_G &= \log p^t(G^\prime | G)
\end{align*}
Therefore, a valid geometric configuration of target vectors $\mathbf{e}_G$ exists. By the Universal Approximation Theorem for Neural Networks, and since node labels are not repeated (all goals are distinct in the derivation tree), a sufficiently powerful MLP can learn the finite mapping from each unique input identifier $G$ to the corresponding target embeddings $\mathbf{e}_G$ with arbitrary precision.
% This completes the proof.
\end{proof}

\subsection{Derivation of Recursive Value Function}
\label{app:valuefunc}

Following the standard MDP formulation,
\begin{align*}
V^{\pi_\lambda}(G, y) 
&= \sum_{a \in \mathcal{A}(G)} \pi_\lambda(\operatorname{res}(G, a), G) \sum_{G'} P(G' \mid G, a) \cdot \\
&\quad \left[ R(G, y) + \gamma V^{\pi_\lambda}(G', y) \right],
\end{align*}
we simplify this in our SLD-resolution MDP (cf. Section~\ref{sec:mdpmodelling}), where transitions are deterministic
% $
% P(G' \mid G, a) = \mathbf{1}\{G' = \operatorname{res}(G, a)\},
% $
and zero rewards for non-terminal states.
% $
% R(G, y) = 0 \text{ if } G \text{ non-terminal}, V^{\pi_\lambda}(G, y) = R(G, y) \text{ if terminal}. 
% $
Setting \(\gamma = 1\), the value function for non-terminal \(G\) simplifies to 
\begin{align*}
V^{\pi_\lambda}(G, y)
= & \displaystyle \sum_{a \in \mathcal{A}(G)} \pi_\lambda(\operatorname{res}(G, a), G) \cdot\\
& ~~~~~~~~~~~ ~ V^{\pi_\lambda}(\operatorname{res}(G, a), y) \ , \nonumber
\end{align*}
with terminal condition \(V^{\pi_\lambda}(G, y) = R(G, y)\).

\subsection{Alignment between MDP Objective and Cross-entropy Loss}
\label{app:ALIGN}

In NeSy systems, learning is often supervised via cross-entropy between predicted success probabilities and binary labels \cite{winters2022deepstochlog}. We now show that, under our MDP formulation in Section~\ref{sec:mdpmodelling}, the objective that maximizes the expected cumulative reward is equivalent to minimizing cross-entropy loss on a labeled dataset $\mathcal{D} = \{(q^{(i)}, y^{(i)})\}_{i=1}^N$.

\noindent\textbf{NeSy objective.} The NeSy systems minimize the expected cross-entropy loss:
\begin{align*}
\mathcal{L}(\theta) = - \sum_{i=1}^N \Big[ 
& \; y^{(i)} \log p_{\text{success}}^{\lambda}(q^{(i)}) \\
& + (1 - y^{(i)}) \log \left(1 - p_{\text{success}}^{\lambda}(q^{(i)})\right)
\Big].
\end{align*}

\noindent\textbf{MDP objective.} Given the recursive value function from Appendix~\ref{app:valuefunc}, the expected return of a query $q$ with label $y$ can be unrolled as:
{\small
\begin{align*}
V^{\pi_\lambda}(q, y)
&= \sum_{a_0, \dots, a_{T-1}} \prod_{t=0}^{T-1} \pi_\lambda(G_{t+1}, G_t) \cdot V^{\pi_\lambda}(G_T, y),\\
&= \sum_{a_0, \dots, a_{T-1}} \prod_{t=0}^{T-1} \pi_\lambda(G_{t+1}, G_t) \cdot R(G_T, y),
\end{align*}}
where $G_{t+1}=\operatorname{res}(G_t, a_t)$, and we assume the derivation proceeds for $T$ steps, reaching terminal goal $G_T$ at depth $T$. 

From the reward design in Equation~\ref{eq:reward}, only successful derivations that terminate in \textit{True} yield non-zero rewards. By the definition of query success probability in Definition~\ref{def:successSLD}, the value simplifies to: 
{\small
\begin{align*}
V^{\pi_\lambda}(q, y)
&= \sum_{a_0, \dots, a_{T-1}} \prod_{t=0}^{T-1} \pi_\lambda(G_{t+1}, G_t) \cdot R(\textit{True}, y),\\
&= R(\textit{True}, y) \cdot \sum_{d \in \mathcal{R}(q)}  p^{\lambda}(d) 
= (2y-1)p_{\text{success}}^{\lambda}(q).
\end{align*}}

Thus, the MDP objective becomes:
\[
J(\lambda) = \sum_{i=1}^{N} V^{\pi_\lambda}(q^{(i)}, y^{(i)}) = 
\sum_{i=1}^{N} (2y^{(i)} - 1)p_{\text{success}}^{\lambda}(q^{(i)}).
\]
\begin{proof}
We now compare the gradients of the two objectives. Let $P_i := p_{\text{success}}^\lambda(q^{(i)})$, and apply the chain rule:
\begin{align*}
&\nabla_\lambda J(\lambda) = \sum_{i=1}^N (2y^{(i)} - 1) \cdot \nabla_\lambda P_i, \\
&\nabla_\lambda \mathcal{L}(\lambda) = \sum_{i=1}^N \frac{P_i - y^{(i)}}{P_i(1 - P_i)} \cdot \nabla_\lambda P_i.
\end{align*}

Let's analyze the coefficients of $\nabla_\lambda P_i$ in both expressions:
\begin{itemize}
    \item If $y^{(i)} = 1$:
    \begin{itemize}
        \item Coefficient in $\nabla_\lambda J(\lambda)= (2y^{(i)} - 1) = (2 \cdot 1 - 1) = 1$.
        \item Coefficient in $-\nabla_\lambda \mathcal L(\lambda)=\frac{y^{(i)} - P_i}{P_i(1-P_i)}=\frac{1 - P_i}{P_i(1-P_i)} = \frac{1}{P_i}$ (for $P_i \neq 1$).
    \end{itemize}
    Both are positive for $P_i \in (0, 1)$, pushing $\lambda$ towards increasing $P_i$.
    \item If $y^{(i)} = 0$:
    \begin{itemize}
        \item Coefficient in $\nabla_\lambda J(\lambda)= (2y^{(i)} - 1) = (2 \cdot 0 - 1) = -1$.
        \item Coefficient in $-\nabla_\lambda \mathcal L(\lambda)=\frac{y^{(i)} - P_i}{P_i(1-P_i)}=\frac{0 - P_i}{P_i(1-P_i)}  = -\frac{1}{1-P_i}$ (for $P_i \neq 0$).
    \end{itemize}
    Both are negative for $P_i \in (0, 1)$, pushing $\lambda$ towards decreasing $P_i$.
\end{itemize}
Despite the different weighting of the linear combinations of the vectors $\nabla_\lambda P_i$, the coefficient always has the same sign. This confirms that both optimization objectives are fundamentally pushing the parameters $\lambda$ in directions that aim to increase the success probability $P_i$ for positive queries and decrease it for negative queries.

\end{proof}

\section{MDP Implementation}
\subsection{Memory-enhanced Environment}
To enable effective reinforcement learning and prevent loops during derivation, we augment the environment with a memory mechanism that tracks visited goals within a single derivation episode. At each step, the environment excludes any action (clause) that would lead to a previously encountered goal, ensuring that the agent explores only new goals.

\subsection{False Action} \label{app:falseaction}
We extend the action space with a special \textit{False} action, which leads to the \textit{False} terminal goal. This serves two purposes: (i) to ensure the agent remains probabilistic even when only one clause is unifiable, and (ii) to enable learning when to abandon early.

For example, consider a logic program \( \mathcal{P} = \{C_1, C_2, C_3\} \), where $C_1 = \textit{locIn}(X, Y) \leftarrow \textit{neighOf}(X, Z), \textit{locIn}(Z, Y)$, $C_2 = {} \leftarrow \textit{neighOf}(\textit{tr}, \textit{gr})$,  
and $C_3 =  {}\leftarrow \textit{locIn}(\textit{gr}, \textit{eu})$. Consider a negative query \( q = G_0 = {}\leftarrow \textit{locIn}(\textit{tr}, \textit{eu}) \) with label \( y = 0 \). The only unifiable clause is $C_1$, resulting in a deterministic step. To preserve stochastic behavior, we add \textit{False} action to the action space: \( \mathcal{A}(G_0) = \{(C_1, \{X/\textit{tr}, Z/\textit{eu}\}), \textit{False}\} \). This allows the agent to distribute probability mass across multiple actions, including the option to terminate early by selecting \textit{False}.

\section{Experiments}

To ensure reproducibility across all experiments, we set the random seeds for all relevant libraries and components. Specifically, we initialize the seed for Python's built-in random module, NumPy, and PyTorch. For GPU computations, we set the seed for CUDA’s random number generators on all devices. Additionally, we configure PyTorch’s backend settings by enabling deterministic operations and disabling benchmarking to prevent nondeterministic optimizations during training.

\subsection{MNIST Addition}
\label{app:mnist}
\paragraph{Dataset.}
We follow the same data generation process for the MNIST addition experiment as in previous works \cite{manhaeve2018deepproblog}. This involves consecutively selecting \(2N\) images from the MNIST dataset and splitting them into two distinct sequences, each of length \(N\). To create supervision signals, we compute the target sum values by multiplying the labels of the selected MNIST images by the corresponding powers of 10 and summing the resulting sequences. Finally, the two numbers are added together to form the final label. Each MNIST image is used only once in all sequences, resulting in \(\lfloor 60000 / 2N \rfloor\) learning samples. They are further split in a 5:1 ratio for training and validation. The test set is generated similarly using the MNIST test partition.

\paragraph{Neural Network.} As in previous works, we use a standard LeNet architecture in all MNIST addition experiments to output the probability distribution over digits, indicating the likelihood that each image in the two sequences corresponds to a specific digit \cite{manhaeve2018deepproblog}.

\paragraph{Dynamic Programming (DP).} 
We adopt a structured factorization of the problem as digit-wise addition with carry propagation, similar to DeepSoftLog~\cite{maene2023soft}. This formulation scales linearly in the sequence length $N$. During DP, we maintain two tables: one indexed by $(i, s_i)$ and another by $(i, c_i)$, where $i \in \{0, \ldots, N-1\}$ denotes the digit position (with $i = 0$ being the least significant digit), $s_i$ is the observed sum digit at position $i$, and $c_i$ is the carry propagated from position $i$ to $i+1$. Each entry in the DP tables represents the total probability mass of all valid latent explanations (i.e., digit assignments) leading to that state. The probability computation accounts for the distribution over the incoming carry from the previous digit, as well as the output distributions of the neural digit classifier for the two MNIST images at the current position. Since we train only on positive queries in this problem, maximizing the value function corresponds to maximizing the probability of the ground-truth sum. This probability is computed as the product of the probabilities assigned to the correct sum digit at each position.

The recursive structure of MNIST addition can be represented by the following logic program:

\begin{lstlisting}[
  breaklines=true,
  breakindent=0pt,
  breakautoindent=false,
  aboveskip=0.5em,
  belowskip=0.5em
]
mnist_addition([], [], [], 0).
mnist_addition([], [], [1], 1).
mnist_addition([HN1|TN1], [HN2|TN2],[HSUM|TSUM], CarryIn) :- 
    Sum is HN1+HN2+CarryIn, 
    HSUM is Sum mod 10, 
    CarryOut is Sum // 10, 
    mnist_addition(TN1, TN2, TSUM, CarryOut).
\end{lstlisting}

In the recursive clause, the head represents the current goal. For a given digit pair (100 combinations in total), the only non-deterministic atom in the body is the recursive call advancing to the next digit position. As a result, the goal space grows linearly with $N$. This structure provides a general insight for neurosymbolic problem representations: if a problem can be encoded as a recursive formula in which the only non-deterministic atom is the next goal and the number of substitutions is tractable, then the problem can be solved efficiently using DP.

\paragraph{Policy Gradient.} The logic program for masking out invalid actions is shown below. In each training iteration, we sample rollouts that satisfy the ground-truth sum from the masked distribution. These rollouts are used to update the unmasked policy in an off-policy manner, with importance sampling applied to correct for the distribution mismatch.

\begin{lstlisting}[
  breaklines=true,
  breakindent=0pt,
  % breakautoindent=false,
  aboveskip=0.5em,
  belowskip=0.5em
]
% Generate a mask list of 10 entries (0-9)
digit_mask(Pos, SeqLen, SumDigit, FullSum, CurrNo, Prev, Mask) :-
    findall(Flag, (
        between(0, 9, D),
        ( valid_digit(Pos, SeqLen, SumDigit, FullSum, CurrNo, Prev, D) ->
            Flag = 1
        ;
            Flag = 0
        )
    ), Mask).

% Case: last position, first number
valid_digit(Pos, SeqLen, SumDigit, _, 0, _, D) :-
    Pos =:= SeqLen - 1,
    Remain is SumDigit - D,
    between(0, 9, Remain).

% Case: not last, first number
valid_digit(Pos, SeqLen, SumDigit, FullSum, 0, _, D) :-
    Pos < SeqLen - 1,
    Remain is SumDigit - D,
    RemainLen is SeqLen - Pos - 1,
    max_suffix(FullSum, RemainLen, MaxSuffix),
    (MaxSuffix -> MaxDiff = 9 ; MaxDiff = 10),
    between(0, MaxDiff, Remain).
    
% Case: last position, second number
valid_digit(Pos, SeqLen, SumDigit, _, 1, Prev, D) :-
    Pos =:= SeqLen - 1,
    Pred is Prev + D,
    Pred =:= SumDigit.

% Case: not last, second number
valid_digit(Pos, SeqLen, SumDigit, FullSum, 1, Prev, D) :-
    Pos < SeqLen - 1,
    Pred is Prev + D,
    RemainLen is SeqLen - Pos - 1,
    max_suffix(FullSum, RemainLen, MaxSuffix),
    ( MaxSuffix -> Pred =:= SumDigit ; PredSum =:= SumTarget - 1 ).

% Helper predicates
max_suffix(FullSum, RemainLen, true) :-
    pow10(R, RemainLen),
    FullSum mod R =:= R - 1, !.
max_suffix(_, _, false).
pow10(1, 0).
pow10(Result, N) :-
    N > 0,
    N1 is N - 1,
    pow10(R1, N1),
    Result is R1 * 10.


\end{lstlisting}

\paragraph{Computation Resources.} We conduct our experiments on a workstation equipped with 2*AMD EPYC 7502 CPUs @ 2.5 GHz, 256 GB RAM, and 4*NVIDIA RTX A5000 GPUs (24GB each). All experiments are executed on a single GPU. Since training time can be significantly affected by the current load on the workstation, we report a normalized training time for fair comparison. Specifically, we estimate total training time by measuring the duration of a single epoch on a laptop with an Intel Core i7-10850H CPU @ 2.70GHz, 32GB RAM, and an NVIDIA GeForce RTX 2070 with Max-Q Design (8GB), and multiplying it by the number of epochs required for convergence on the workstation.

\paragraph{Hyperparameters}
The optimal hyperparameters for \acr{}(DP), \acr{}(PG), and Scallop were determined via grid search on a held-out validation set. We evaluated learning rates \(\{0.0003, 0.001, 0.003\}\) and batch sizes \(\{32, 64, 128\}\). For \acr{}(DP), the maximum number of training epochs was set to 5000 for all values of \(N\). The best results for \(N = 4, 15, 200\) were achieved with a learning rate of 0.003 and batch size 128, while for \(N = 100 \text{ and } 500\), the optimal configuration was a learning rate of 0.001 and batch size 64. For \acr{}(PG), the maximum number of training epochs was set to 10,000 for all \(N\). Experiments with \(N = 4, 15\) performed best using a learning rate of 0.001 and batch size 32. For \(N = 100, 200, 500\), we report performance using a learning rate of 0.003 and batch size 32. For Scallop, the maximum number of training epochs was set at 20 across all \(N\). Due to its top-\(k\) approximate inference, selecting \(k\) is critical to balance accuracy and computational efficiency. We explored \(k \in \{1, 3, 5, 6, 7\}\) based on the original Scallop implementation. In all cases where Scallop did not time out, \(k=3\) with a learning rate of 0.001 and batch size 32 yielded state-of-the-art performance and the fastest runtimes. Each experiment was run with 5 different random seeds.

For KLay \cite{DBLP:conf/iclr/MaeneDM25}, DeepStochLog \cite{winters2022deepstochlog}, DeepSoftLog \cite{maene2023soft}, A-NeSI \cite{van2023nesi}, and EXAL \cite{verreet2024explain}, we used their best reported accuracies. The training times were obtained with the optimal hyperparameters provided in their respective papers.

\subsection{Explainable Knowledge Graph Completion}
\label{app:KG}

\paragraph{Detailed dataset statistics} The statistics for the datasets used are provided in Table 4.

\begin{table}[h!]
\centering
\footnotesize
\begin{tabular*}{\columnwidth}{@{\extracolsep{\fill}} l r r r r r @{}}
\toprule
Dataset & Entities & Rels. & Facts & Avg. Degree & Rules \\
\midrule
Family  & 3,007  & 12 & 19,845 & 6.47 & 48 \\
WN18RR  & 40,559 & 11 & 86,835 & 2.14 & 28 \\
\bottomrule
\end{tabular*}
\caption{Detailed dataset statistics.}
\label{tab:final-table}
\end{table}

\paragraph{Evaluation Metrics for KG Completion}
To assess model performance in the link prediction task, we employ three standard ranking metrics:

Mean Reciprocal Rank (MRR). This metric calculates the average reciprocal of the rank of the first correct entity for a set of test queries $Q$:
$$ \text{MRR} = \frac{1}{|Q|} \sum_{q \in Q} \frac{1}{\text{rank}_q} $$
where $\text{rank}_q$ is the rank of the correct entity for query $q$. 

Hits@N. This metric measures the proportion of queries for which a correct entity appears in the top-$N$ ranked predictions. It is calculated as:
$$ \text{Hits@N} = \frac{1}{|Q|} \sum_{q \in Q} \mathbb{I}(\text{rank}_q \le N) $$
where $\mathbb{I}(\cdot)$ is the indicator function.

Area Under the Precision-Recall Curve (AUC-PR). This metric computes the area under the Precision-Recall curve, which plots precision against recall at various decision thresholds. AUC-PR is particularly informative for imbalanced datasets, common in link prediction, as it focuses on the performance of positive instances.

\paragraph{Computation Resources.}
We used an Intel Core i7 CPU @ 2.10 GHz, 48 GB of RAM, 2*NVIDIA Tesla V100-SXM2 GPUs (32 GB each).

\paragraph{Hyperparameters.}
For the knowledge graph experiments, due to the cost of running multiple seeds, the seed was fixed to 0.
We explored the values of 50-200 for embeddings size, {0.2, 0.5} for entropy coefficient, and {0.1, 0.2} for clip range. For both datasets, the entropy coefficient of PPO was set to 0.2, as well as the clip range. The learning rate is set to 3e-4 and the embedding size 64. The rest of the parameters can be consulted in the repository.

\end{document}